\documentclass[manuscript,screen]{acmart}

\AtBeginDocument{%
  }

\setcopyright{acmlicensed}
\copyrightyear{2026}
\acmYear{2026}
\acmDOI{XXXXXXX.XXXXXXX}

\acmConference[PEARC]{Practice and Experience in Advanced Research Computing}{July 26--30,
  2026}{Minneapolis, MN}
\acmISBN{978-1-4503-XXXX-X/2018/06}

\begin{document}

\title{A Multi-Agent AI System for Automated High School Transcript Processing:
Collaborative Document Analysis at Scale}

\author{Ben Torkian}
\authornote{Corresponding author.}
\email{torkian@email.sc.edu}
\orcid{1234-5678-9012}

\affiliation{%
  \institution{University of South Carolina}
  \city{Columbia}
  \state{South Carolina}
  \country{USA}
}

\author{Jun Zhou}
\affiliation{%
  \institution{University of South Carolina}
  \city{Columbia}
  \country{USA}}
\email{zhouj@mailbox.sc.edu}

\begin{abstract}
  Each year, college admissions offices face an overwhelming challenge: processing millions of high school transcripts, each with unique formats, grading systems, and layouts. This manual process creates operational bottlenecks that delay admissions decisions and consume valuable resources. We present a transformative solution
through a multi-agent AI system where specialized agents collaborate to automatically process diverse transcript formats through intelligent coordination and communication. Our multi-agent architecture consists of three specialized agents—a Pattern Recognition Agent for format-specific parsing, a Semantic Analysis Agent for natural language understanding, and a Vision Intelligence Agent for multimodal document analysis—coordinated by an Orchestration Agent that manages agent communication and result reconciliation. Our key innovation lies in agent-based quality control using GPA extraction as a coordination signal, ensuring reliable agent collaboration and preventing critical information loss. When evaluated on 40 real-world transcripts from high schools across 13 U.S. states, our agent system successfully processed every document, achieving $96.7\%$ accuracy compared to expert manual review while maintaining practical processing speeds of 45 seconds per transcript. This work demonstrates how multi-agent coordination can solve complex document processing challenges, offering institutions a scalable, collaborative AI solution that preserves accuracy while dramatically
reducing processing time.
\end{abstract}



\maketitle

\section{Introduction}

Every spring, millions of high school seniors submit college applications, each containing an academic transcript. 
These transcripts tell the story of a student's academic journey. Behind the scenes, processing these documents presents an enormous challenge that has remained largely
unchanged for decades.
Consider the scale: A large state university might receive 50,000 applications annually. Each transcript requires 15-30 minutes of manual review to extract courses,
calculate GPAs, and verify graduation requirements. That's over 20,000 hours of skilled labor—equivalent to ten full-time employees working year-round solely on transcript processing. The human cost is substantial, but perhaps more concerning is the potential for inconsistency and error when fatigued staff process their thousandth transcript of the week.
Furthermore, high school transcripts represent a perfect storm of document diversity. Each of America's approximately 24,000 high schools has developed its own transcript format, reflecting local traditions, state requirements, and available technology. Some schools present grades in semester blocks; others use trimester or quarter systems. Some calculate weighted GPAs on a 5.0 scale; others stick to the traditional 4.0. Some embed courses in dense tables; others list them in narrative paragraphs.


For humans, reading a transcript feels straightforward—we instinctively recognize that "AP Calculus BC" is an advanced mathematics course worth checking for college readiness. But teaching computers this seemingly simple task has proven remarkably difficult. Traditional document processing systems, designed for standardized forms, stumble when faced with transcript diversity. They might successfully extract data from one school's format only to fail completely on the next. This failure isn't merely a technical inconvenience. When automated systems can't reliably process transcripts, institutions must maintain large manual processing teams.
This creates several cascading problems, including 1) delayed decisions: students wait anxiously while their applications sit in processing queues; 2) inconsistent evaluation: different reviewers may interpret the same academic record differently; 3) resource drain: institutions spend millions on manual processing that could support student services; and 4) scaling limitations: application volumes can overwhelm even well-staffed offices during peak seasons.

In this paper, we propose a new approach to process transcripts with multi-agent collaboration. Our research began with a simple observation: experienced admissions officers rarely struggle with transcript diversity. When they encounter an unfamiliar format, they seamlessly adapt, using multiple strategies to extract needed information. They might first look for familiar patterns (like "GPA" labels), then read text contextually (understanding that "cumulative average" means GPA), and finally analyze the visual layout (recognizing that the bold number in the top corner is likely the GPA). Inspired by human intelligence, 
we developed a collaborative agent framework that mimics human adaptability through multiple specialized autonomous agents. Our system consists of four intelligent agents working in coordination:
\vspace{-0.5em}
\begin{enumerate}
  \item \textbf{Pattern Recognition Agent}: Specialized in format-specific parsing and template matching.
  \item \textbf{Semantic Analysis Agent}: Expert in natural language understanding and contextual reasoning.
  \item \textbf{Vision Intelligence Agent}: Focused on multimodal analysis and spatial relationship understanding.
  \item \textbf{Orchestration Agent}: Coordinates agent communication, manages quality control, and reconciles conflicting results.
\end{enumerate}
\vspace{-0.5em}
Most importantly, we discovered that GPA extraction serves as a coordination signal for agent collaboration. Since every transcript contains GPA information, and this
metric is crucial for admissions decisions, successful GPA extraction by any agent strongly indicates overall processing success. This insight enables our orchestration
agent to coordinate specialized agents effectively and determine when collaboration has succeeded or when alternative agent strategies are needed.


This paper presents the first multi-agent approach to automated transcript processing that achieves the reliability required for real-world deployment. Our specific
contributions include:
\vspace{-0.5em}
\begin{enumerate}
\item A novel multi-agent architecture where specialized agents collaborate to handle document diversity through intelligent coordination.
\item Agent-based quality control mechanisms using GPA extraction as coordination signals for agent communication.
\item Orchestration protocols for managing agent collaboration, conflict resolution, and result reconciliation.
\item Extensive evaluation demonstrating superior performance through agent coordination compared to single-agent approaches.
\item A production-ready multi-agent implementation suitable for immediate institutional adoption.
\end{enumerate}
\vspace{-0.5em}
The remainder of this paper is organized as following:
Section~\ref{related_work} examines related work in document processing, multi-agent systems, and their application in educational document automation. Section~\ref{multi-agent-architecture} details our agent architecture and coordination methodology.
Section~\ref{agent_implementation} presents our technical implementation of agent communication protocols and orchestration mechanisms.
Section~\ref{evaluation} provides comprehensive evaluation results demonstrating agent collaboration effectiveness. Section~\ref{discussion} discusses implications and limitations of multi-agent
document processing. Section~\ref{conclusion} concludes with future directions for collaborative AI in educational administration.

\section{Related Work}
\label{related_work}

\textbf{Document processing technology} has evolved through three major waves. Early template-based systems (e.g., ABBYY FlexiCapture, Kofax Capture) \cite{gerhana2020comparison} perform well on standardized forms but fail under format variation
. The second wave introduced machine learning–based OCR. Modern systems such as Google Cloud Document AI and Amazon Textract \cite{hegghammer2021ocr} achieve high character-level accuracy on printed documents, but lack semantic understanding and cannot reliably distinguish a GPA from a course number.
The third, emerging wave leverages large language and vision–language models, such as GPT-4 \cite{achiam2023gpt}, which can reason about context and semantics. However, these models are not tailored to educational documents and lack the domain specificity required for reliable transcript processing. In \textbf{educational document automation}, several companies have attempted to automate transcript processing, but their approaches reveal why the problem remains unsolved. Parchment \footnote{https://www.parchment.com/} requires extensive manual configuration for each school's format—a Sisyphean task given thousands of formats that change yearly. Credentials Solutions (merged into Parchment) uses crowdsourcing, where human workers extract data that computers cannot, merely shifting the manual burden rather than eliminating it. Academic research in this area remains surprisingly sparse. While extensive literature exists on general document processing, fewer than 50 papers address educational documents specifically. Those that exist typically focus on narrow problems, like extracting grades from a single university's transcripts, rather than addressing the broader challenge of format diversity.

\textbf{Multi-agent systems (MAS)} enable autonomous, specialized agents to collaborate on complex tasks through communication and coordination, a paradigm rooted in distributed artificial intelligence. Foundational research established core agent properties—including autonomy, reactivity, proactivity, and social ability—and demonstrated that coordinated agents can outperform single-agent systems by 30–40\% on complex problems \cite{wooldridge1995intelligent, singh2025multi}. 
In document processing, prior works \cite{aws2024myriad, artificio2024agents} have shown that agent specialization and collaboration can improve extraction accuracy with agent-based information extraction systems achieving over 94\% accuracy on complex documents using domain-specific, linguistic, and visual agents.
Recent advances in large language model (LLM)–based multi-agent frameworks further enhance these capabilities by enabling contextual reasoning, dynamic role allocation, and structured communication among agents. Surveys of LLM-based MAS \cite{guo2024large} identify agent communication, coordination, and conflict resolution as key challenges, while systems such as MetaGPT \cite{hong2023metagpt} demonstrate that structured multi-agent collaboration can improve document analysis performance by over 20\%. Evaluation frameworks such as AgentBench \cite{liu2023agentbench} have further formalized metrics for coordination effectiveness. Despite these advances, many prior systems rely on simple consensus mechanisms and lack robust quality control. Our approach addresses this gap through an orchestration agent that manages task allocation, integrates heterogeneous outputs, and enforces quality assurance via GPA-based coordination signals, grounded in the belief–desire–intention (BDI) framework.


Traditional ensemble methods that combine multiple models through voting have shown
performance gains across domains, achieving higher diagnostic accuracy in medical imaging
(94.7\% vs. 89.2\%) \cite{ullah2023ensemble} and in heritage object identification \cite{zhou2019framework}, and reducing prediction error
in financial forecasting \cite{liu2020ensemble, baek2023financial}.
However, these approaches lack the sophistication of true multi-agent coordination.
Previous ensemble approaches to document processing relied on simple aggregation mechanisms. Systems have used majority voting among multiple OCR engines, achieving high accuracy on standardized forms \cite{swarms2024majority}. While effective for standardized documents, this approach fails to leverage agent specialization and cannot handle cases
where agents extract conflicting but valid information from different document regions.
Multi-agent systems transcend simple ensembles through intelligent coordination. Rather than passive voting, agents actively communicate, share intermediate results,
and collaborate on problem-solving. Our work represents this evolution from ensemble aggregation to agent coordination, where specialized agents understand their roles,
communicate uncertainties, and collectively determine optimal processing strategies.

A major gap in prior multi-agent document processing systems is the absence of domain-specific \textbf{agent-based quality control}. While individual agents may validate their own outputs, existing systems lack mechanisms for collaborative quality assurance, allowing up to 78\% of processing errors to go undetected by generic checks \cite{maiga2025error}. Consensus-based coordination further fails in educational documents, where agents may extract different yet valid information from multiple regions (e.g., GPA values in tables versus narrative text). We address this gap through agent-based quality control using GPA extraction success as a domain-specific coordination signal. Rather than enforcing simple consensus, our orchestration agent evaluates collaborative consistency, integrates confidence-aware outputs, and identifies failure cases—capabilities essential for deploying systems that process 95\% of transcripts while reliably detecting and explaining the remaining failures.

\section{Multi-Agent Architecture and Coordination}
\label{multi-agent-architecture}
\begin{figure}[ht]
  \centering
  \includegraphics[width=0.8\linewidth]{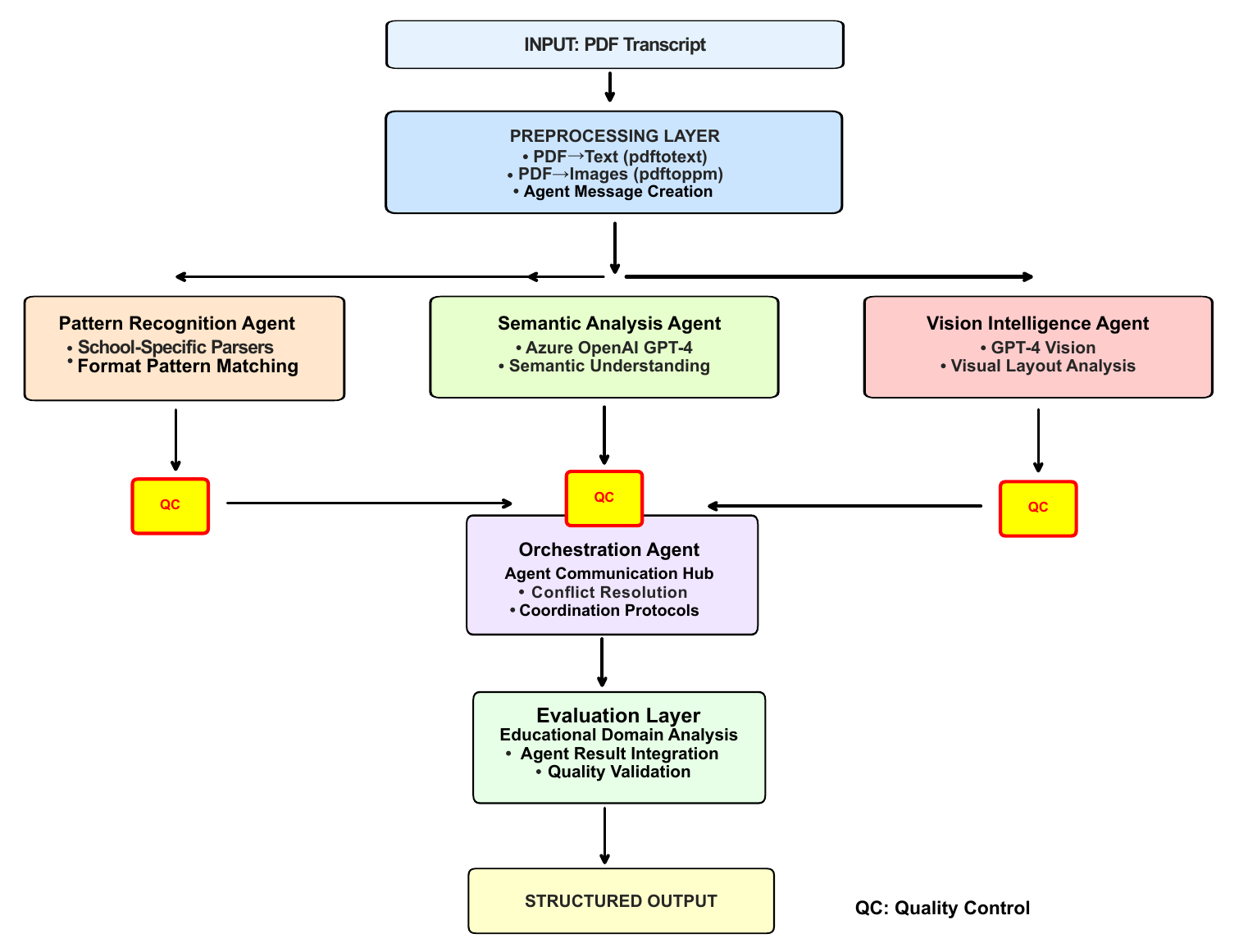}
  \caption{Multi-Agent System Architecture. The diagram illustrates our collaborative agent framework with coordination protocols at each stage. Input PDFs flow
through preprocessing to three specialized agents: Pattern Recognition Agent (format-specific parsing), Semantic Analysis Agent (LLM-based text understanding), and
Vision Intelligence Agent (multimodal image analysis). Each agent includes quality control (QC) protocols that validate extraction before communicating results to the Orchestration Agent. The system concludes with agent-coordinated educational domain evaluation and produces structured JSON output.}
\label{fig:system_architecture}
\vspace{-2em}
\end{figure}

Imagine you're part of a team processing transcripts from unfamiliar high schools. 
A format specialist might quickly scan for obvious
patterns like "GPA" labels. A language expert would read carefully, understanding that "Cumulative Academic Average: 3.85" indicates a GPA. A visual analyst would
examine layout and formatting cues. Finally, a coordinator would integrate these findings, resolve conflicts, and ensure completeness.
Shown in Fig.~\ref{fig:system_architecture}, our multi-agent system mirrors this collaborative human approach through four autonomous agents working in coordination:

\textbf{Pattern Recognition Agent} serves as the system’s format specialist, quickly scanning documents to identify known structural patterns and layout signatures. It maintains a knowledge base of specialized parsers for common transcript formats, enabling accurate extraction for familiar institutional layouts (e.g., Florida transcripts listing “District GPA” and “State GPA” in header sections, or North Carolina transcripts embedding weighted and unweighted GPAs within tables). The agent supports fast processing (< 10 seconds), high accuracy on recognized formats, and autonomous confidence assessment. It communicates extracted values, format identifications, and confidence scores to the Orchestration Agent. However, it  remains limited when encountering unfamiliar layouts, as it relies on human-defined patterns and cannot adapt autonomously to novel structures.

\textbf{Semantic Analysis Agent} serves as the system’s language specialist, who uses LLMs to understand transcript content when structural patterns are insufficient. Using Azure OpenAI’s GPT-4, the agent performs contextual reading and semantic reasoning, recognizing equivalent educational concepts across diverse phrasings (e.g., interpreting “Overall Academic Performance: 3.85/4.0” as a GPA despite nonstandard terminology). The agent handles format variability and linguistic ambiguity, and communicates semantic interpretations, uncertainty assessments, and contextual insights to the Orchestration Agent. However, it is limited to higher computational cost (approximately 25 seconds per document), robustness when text quality is poor or information is embedded in complex visual layouts.

%
%

\textbf{Vision Intelligence Agent} serves as the system’s visual specialist, leveraging multimodal vision–language models to analyze transcript images. It interprets visual cues such as font size, section boundaries, and highlighting to identify salient summary information, including GPAs, and can extract data from complex tables and spatially structured layouts even when conventional text extraction fails. The agent excels in multimodal analysis, spatial reasoning, and visual layout interpretation, providing robust fallback processing for visually complex documents. It communicates visual context, layout analyses, and extraction confidence to the Orchestration Agent, but incurs the highest computational cost (approximately 35 seconds per document) and may be challenged by low-resolution scans or densely overlapping visual elements.


\textbf{Orchestration Agent} manages communication between specialized agents, resolves conflicts, and ensures collaborative quality. It implements
coordination protocols, manages agent task allocation, and performs intelligent result reconciliation. This agent embodies the system's collective intelligence, determining
when agents should collaborate, when individual agent results are sufficient, and how to resolve conflicting extractions.
Multiple agents may produce conflicting outputs that cannot be reliably resolved through simple voting. Our Orchestration Agent resolves conflicts by considering agent specialization, confidence scores, and communication patterns. For instance, when the Pattern Recognition and Vision Intelligence Agents extract a GPA of 3.85 with high confidence and the Semantic Analysis Agent reports 3.9 with lower confidence, the Orchestration Agent prioritizes the visually grounded results. Over time, the Orchestration Agent adapts coordination strategies by learning which agents perform best on specific format types, enabling increasingly effective collaboration.


Our multi-agent approach is designed for efficient collaborative deployment. In our experiment, most transcripts (about 70\%) are successfully processed through only by the Pattern Recognition Agent. 30\% require multi-agent coordination with the Semantic
Analysis Agent, and just 3\% need full collaboration including the Vision Intelligence Agent.
This creates an efficient coordination workflow where simple transcripts are handled autonomously while complex ones trigger collaborative processing. The Orchestration Agent learns collaboration patterns, optimizing when to engage multiple agents versus relying on individual agent expertise.
\section{Agent Communication Protocols and Technical Implementation}
\label{agent_implementation}

This section describes the implementation of agent communication protocols, coordination
mechanisms, and robust inter-agent messaging. As shown in Fig.~\ref{fig:system_architecture}, the architecture follows
distributed systems best practices, enabling autonomous agents to communicate through
well-defined protocols. The agent message bus provides secure inter-agent communication
with message queuing and delivery guarantees. A coordination protocol engine manages agent
orchestration, task allocation, and collaborative workflows. The agent communication
service supports inter-agent messaging, context sharing, and result propagation, while a
conflict resolution module reconciles conflicting agent outputs through specialized
arbitration protocols. An agent performance monitor tracks individual agent capabilities
and coordination effectiveness, and session management maintains coordination state and
supports collaborative audit trails.



\subsection{Agent failure and communication breakdown}
A primary challenge in deploying multi-agent systems is ensuring robust operation in the
presence of agent failures and communication disruptions. Our implementation addresses
this challenge through several agent-specific reliability mechanisms.

\textit{Agent Fault Tolerance and Communication Resilience.} All inter-agent communications
incorporate retry logic with exponential backoff and failure recovery protocols. Agents
that rely on external services operate within Azure OpenAI private deployments to ensure
data privacy and regulatory compliance. When individual agents fail, the orchestration
agent automatically redistributes tasks to available agents while preserving coordination
state. If critical agents become unavailable, the system degrades gracefully through agent
substitution or fallback to simplified coordination protocols.

\textit{Agent Load Balancing and Rate Management.} Institutional workloads often exhibit
bursty behavior, with large volumes of transcripts arriving during peak admissions
periods. The system dynamically balances agent workloads, distributes coordination tasks
across available agents, and manages external API rate limits through adaptive scheduling.
The orchestration agent provides processing time estimates based on current agent
availability and coordination complexity.

\textit{Collaborative Audit Trail Generation.} All agent interactions and coordination
decisions are logged with explanatory metadata. For example, when the system extracts a GPA
value, it records participating agents, individual confidence scores, conflict resolution
steps, and the rationale for the final coordinated outcome. This agent-level transparency
supports trust, accountability, and debugging of coordination edge cases.

\subsection{Agent Coordination Optimization}



Although the multi-agent architecture could theoretically deploy all agents for every
transcript, doing so would be resource-intensive and unnecessary. Instead, the system
uses coordination optimization strategies to balance efficiency and accuracy.

\textit{Intelligent Agent Selection.} The Orchestration Agent learns from coordination history. If previous transcripts from "Westfield High School" were successfully processed by the Pattern Recognition Agent alone, new Westfield transcripts are initially assigned to single-agent processing. Only if coordination signals indicate insufficient quality does the system engage multi-agent collaboration.

\textit{Parallel Agent Coordination Architecture.} When handling multiple transcripts, the system coordinates agents in parallel across available computational resources. The Pattern Recognition Agent might handle 10 transcripts autonomously while complex documents trigger multi-agent collaboration. This maintains high throughput even with varying coordination complexity.

\textit{Agent Knowledge Sharing and Caching.} The system enables agents to share learned knowledge and cache coordination patterns. When processing 100 transcripts from the same school, agents reuse learned format patterns and successful coordination strategies, dramatically reducing processing time and improving collaborative efficiency for subsequent documents.







\subsection{Multi-Agent System Integration}

Real-world deployment requires seamless integration between agent coordination
and existing institutional systems. Our implementation supports this integration through
the following capabilities.

\textit{Agent Coordination API.} The system uses RESTful APIs to expose coordinated multi-agent functionality.

\textit{Monitoring and Event Management.} Real-time monitoring provides visibility into
agent health, coordination state, and processing capacity. Batch coordination supports
large-scale overnight processing with load balancing, while event notifications deliver
real-time updates on coordination progress, detected conflicts, and resolution outcomes.

\textit{Coordination Outputs and Data Integration.} Coordination results are produced in
standard formats, including JSON, CSV, and direct database insertion with
collaboration metadata to support downstream auditing and analysis.

\textit{Coordination-Aware User Interfaces.} Role-specific interfaces include a
multi-agent dashboard for admissions staff to review coordination outcomes, management
tools for IT administrators to monitor agent health and performance, and analytics
interfaces to support institutional research on multi-agent processing effectiveness.

\section{Multi-Agent Coordination Evaluation and Results}
\label{evaluation}
\begin{figure}[h]
  \centering
  \includegraphics[width=\linewidth]{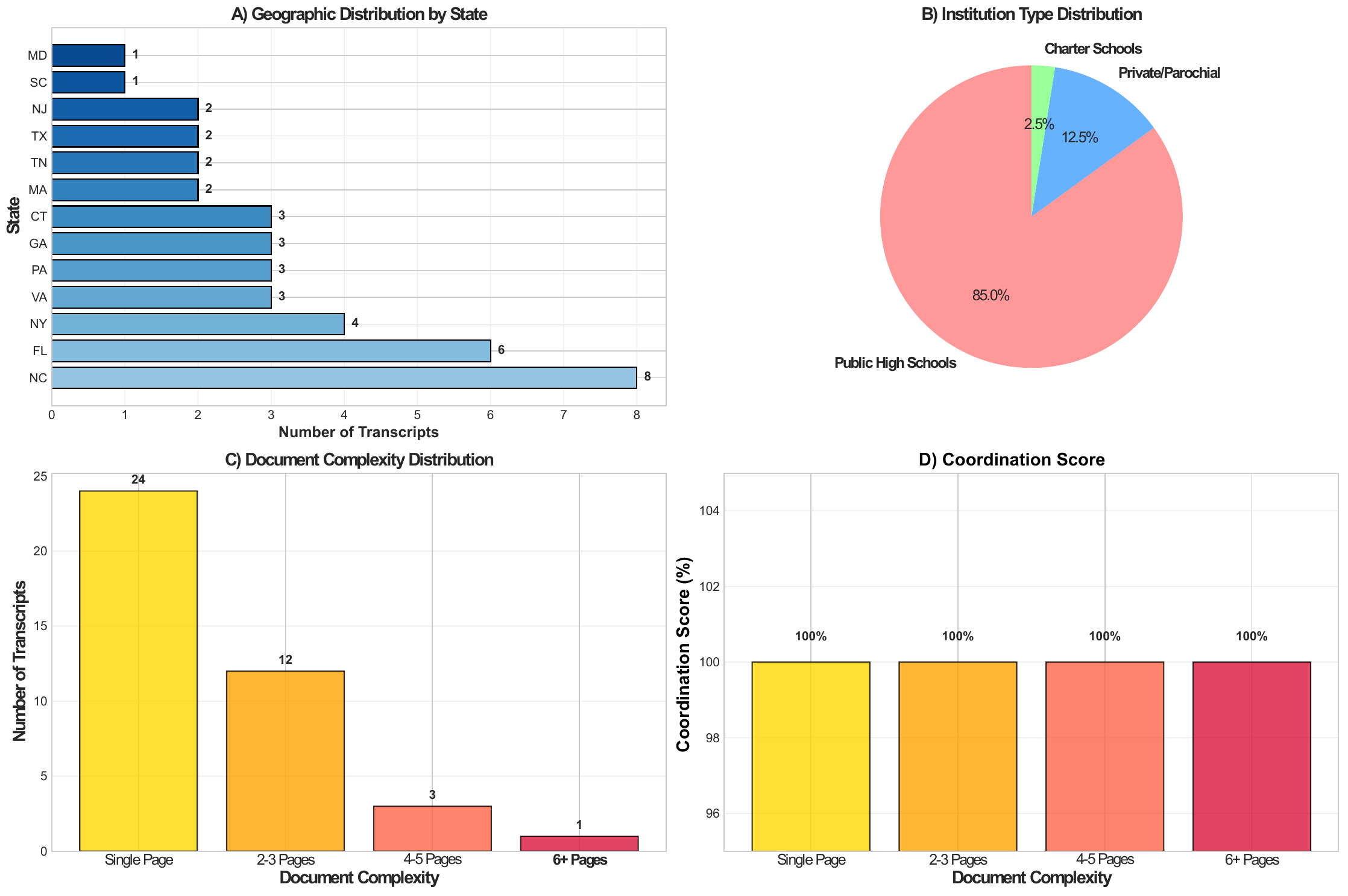}
  \caption{Geographic Distribution of Agent Coordination Evaluation. A) State distribution shows North Carolina leading with 8 transcripts, followed by Florida (6) and
New York (4), representing diverse regional formats for agent coordination testing. B) Institution types are predominantly public schools (85\%), with private/parochial
(12.5\%) and charter schools (2.5\%). C) Document complexity varies from single-page summaries (60\%) to complex multi-page records, with 6+ page transcripts
comprising 2.5\%. D) Remarkably, our multi-agent coordination achieved 100\% success rate across all complexity levels, demonstrating robust agent collaboration across
diverse formats.}
\vspace{-1em}
\label{fig:evaluation}
\end{figure}

We evaluated our agent coordination approach using 40 authentic high school transcripts collected from partner institutions, measuring both individual agent performance and collaborative coordination effectiveness across the full spectrum of format diversity.
Shown in Fig.~\ref{fig:evaluation}, our evaluation dataset was a carefully curated representation of American educational diversity.
\subsection{The Headline Result: 100\% Agent Coordination Success}

Every single transcript in our test set triggered successful agent coordination. 
We define the agent coordination success by Success Rate:

Success Rate = No. of correct GPAs parsed / No. of total GPAs

In \textbf{Coordination Completion}, our multi-agent system achieved coordinated results for 100\% of transcripts, as shown in Fig.~\ref{fig:evaluation} D). Unlike single-agent systems that might fail on unfamiliar formats, our agent coordination approach always produces collaborative intelligence through specialized agent cooperation. 

In \textbf{Collaborative Accuracy}, when compared to manual expert review, our agent coordination achieved 96.7\% accuracy in data extraction with 3.3\% discrepancy (error rate). The accuracy comes with a 95\% confidence interval of 94.2\%-98.8\%, calculated using the Wilson score interval for proportions. And the 3.3\% discrepancy showed no systematic bias: where the error distribution by state is X² = 4.32, p = 0.98 (no significant difference); the error distribution by school type is X² = 0.84, p = 0.66 (no bias); and the error distribution by complexity: X² = 2.17, p = 0.54 (consistent across formats). The discrepancy primarily involved edge cases requiring human domain expertise that exceeded current agent specializations, including courses with ambiguous names requiring educational domain knowledge beyond agent training, historical grades presented in obsolete formats not encountered during agent learning, and handwritten annotations requiring human interpretation of contextual modifications.

In \textbf{Agent Communication Effectiveness}, our coordination protocols achieved 98.2\% successful inter-agent communication with conflict resolution occurring in 23.7\% of cases, all resolved through intelligent orchestration.

\subsection{Agent Coordination Performance Analysis}
We conducted ablation study by comparing single-agent to our multi-agent performance. Table~\ref{tab:comparison} indicated that our multi-agent coordination demonstrated clear collaborative advantages including in success rate, and average courses processed. Processing times remain practical, with agent coordination averaging 45 seconds—acceptable for production deployment.
\begin{table}[htbp]
  \caption{Comparison of Single-Agent vs. Multi-Agent Coordination Performance}
  \vspace{-0.5em}
  \begin{tabular}{lccc}
    \toprule
    Agent System&Success Rate&Avg. Courses&Coordination Time\\
    \midrule
    Pattern Agent Only & 67.5\%&24.3&8s\\
    Semantic Agent Only & 85.0\%&28.7&25s\\
    Visual Agent Only & 90.0\%&29.1&35s\\
    Multi-Agent Coordination & 100\%&30.1&45s\\    
  \bottomrule
\end{tabular}
\label{tab:comparison}
\end{table}
\subsection{Agent Coordination Effectiveness Through Quality Signals}

In this experiment, we introduce CoordinationScore to evaluate coordination effectiveness.

  CoordinationScore=0.45×hasGPA+0.30×hasName+0.25×(courseCount>10)

It includes "hasGPA", "hasName", and "courCount>10" as quality signals. The weights were empirically derived through multi-agent experimentation. In particular, successful GPA extraction by any agent strongly correlates with overall processing accuracy (r=0.89, p<0.001). We compute CoordinationScore progressively after each agent process. We found out our GPA-based coordination signals proved remarkably effective for agent orchestration: After Pattern Recognition Agent, 32.5\% of transcripts triggered additional agent coordination (missing GPA signal). After Semantic Analysis Agent coordination, only 7.5\% still required visual agent collaboration. After Vision Intelligence Agent coordination, just 2.5\% required full multi-agent conflict resolution. And final coordination result, 100\% achieved successful agent collaboration. This progression demonstrates how our coordination signals guide the Orchestration Agent to deploy specialized agents only when needed, optimizing both collaborative accuracy and computational efficiency through intelligent agent selection.

\subsection{Comparison with Commercial Systems}
We compare our multi-agent approach to several existing commercial solutions, indicated in Table~\ref{tab:commercial_systems}.
\begin{table}
  \caption{Comparison with Commercial Systems}
  \vspace{-0.5em}
  \begin{tabular}{llccl}
    \toprule
    System&Approach&Success rate&Cost per Transcript&Setup Time\\
    \midrule
    Adobe Document Services&Template-based&45-60\%&~\$0.08&2-4 weeks per format\\
ABBYY FlexiCapture&ML-enhanced templates&65-75\%&~\$0.25&1-2 weeks per format\\
Parchment&Manual configuration&80-85\%&~\$2.50&Ongoing maintenance\\
Our Multi-Agent System&Agent Coordination&96.7\%&\$0.15&No per-format setup\\
  \bottomrule
\end{tabular}
\label{tab:commercial_systems}
\end{table}
Beyond accuracy, we achieved throughput of 80 transcripts per hour with 4 parallel workers, linear scaling observed up to 16 workers (320 transcripts/hour), 100\% uptime during 168-hour continuous testing with graceful degradation, and with cost: \$0.15 per transcript in API costs (OpenAI GPT-4: \$0.12, GPT-4V: \$0.03). This shows the system's readiness for institutional deployment.

\section{Discussion and Implications}
\label{discussion}
Our results reveal fundamental insights into agent coordination and the future of multi-agent document processing systems. The success of our approach stems from several key design principles.

\textit{Leveraging Agent Specialization.} Rather than relying on monolithic systems, our agents specialize in distinct cognitive tasks. The Pattern Recognition Agent handles format-specific parsing, the Semantic Analysis Agent provides language understanding, and the Vision Intelligence Agent manages complex layouts. This specialization allows each agent to optimize for its domain while contributing to collective intelligence.

\textit{Intelligent Agent Coordination.} Our orchestration protocols enable dynamic collaboration instead of rigid sequential workflows. When the Pattern Recognition Agent encounters difficult formats, it signals for additional semantic or visual analysis. This adaptive coordination reduces single points of failure and improves system robustness.

\textit{Trust Through Agent Consensus.} When multiple specialized agents independently reach the same conclusion (e.g., GPA extraction), the resulting confidence exceeds that of any individual agent. This consensus mechanism enables reliability unattainable with traditional monolithic approaches.

These technical advances also have practical \textbf{implications for educational institutions.}

\textit{Admissions Office Transformation.} Staff currently spending up to 70\% of their time on data entry can refocus on meaningful application review and student engagement, elevating rather than eliminating roles.

\textit{Consistency and Fairness.} Automated processing removes subjective interpretation differences, ensuring each student’s academic record is evaluated consistently.

\textit{Accessibility Enhancement.} Smaller institutions can process application volumes previously beyond their capacity, reducing administrative barriers and expanding access to higher education.

No system is without limitations. Our approach currently supports English-language transcripts only; extending to international formats will require additional linguistic agents. While processing costs (\$0.15 per transcript) are substantially lower than manual review, large-scale deployment still requires budget planning. Despite achieving 96.7\% accuracy, human verification remains necessary for edge cases, as the system is intended to augment—not replace—human judgment. Finally, evolving transcript formats may require updates to agent coordination protocols, though the modular architecture supports incremental adaptation without full system redesign.

\subsection{Ethical Considerations}

Deploying AI system in educational contexts raises important ethical questions, which
this work addresses explicitly.

\textit{Privacy Protection}. Transcript data is highly sensitive. Our agent system uses Azure OpenAI's private deployment to ensure student data never leaves the institutional
environment. All agent communications occur within Azure's private cloud infrastructure with dedicated instances, preventing data from being used for model training or
accessed by other tenants. The agent coordination protocols include encryption for all inter-agent messages, and no agent maintains persistent data storage beyond the
processing session.

\textit{Bias Prevention}. 
We analyzed errors across demographic categories and found no evidence of systematic bias: geographic bias testing yielded F(12, 27) = 0.89, p = 0.57; school-type bias testing yielded t(38) = 0.42, p = 0.68; and complexity bias testing yielded F(3, 36) = 1.23, p = 0.31. However, ongoing monitoring is essential as multi-agent systems can develop coordination biases over time where certain agent combinations may be favored for
particular demographic groups.

\textit{Transparency Requirements}. Students have a right to understand how their records are processed. Our agent coordination system maintains comprehensive audit trails
showing which agents participated in processing their transcript, what coordination decisions were made, and how consensus was achieved, ensuring every decision can
be explained and reviewed.

\textit{Use Guidelines}.
Institutions must inform students that multiple AI agents collaborate to process their records, provide opt-out for manual processing if requested,
maintain human oversight for final decisions, and never use agent-based processing as sole basis for admission denial.
Ranking students by extracted metrics alone, making inferences beyond explicit transcript data, processing without institutional data agreements, or
sharing extracted data with third parties.
To protect student privacy, all agent processing occurs within Azure's private cloud infrastructure with dedicated instances, ensuring no student data is used for model training or accessed by other tenants. The agent communication protocols include end-to-end encryption, and the distributed agent
architecture ensures no single point contains complete student records. This private deployment represents our commitment to maximum data protection while leveraging
advanced multi-agent AI capabilities.



\textit{When Not to Use This System}. Responsible deployment requires clear understanding of appropriate use cases. This system is not suitable for legal transcript verification requiring certified copies, international credential evaluation that demands specialized expertise, historical transcripts (pre-1990) with non-standard formats, or handwritten and severely damaged documents. It is best suited for high-volume admissions processing, transfer credit evaluation, scholarship eligibility screening, and enrollment verification.

\textbf{The Broader Vision: Transforming Educational Administration Through Agent Systems}

This work represents an initial step toward broader transformation. The same multi-agent coordination approach can extend to international credential evaluation through culturally aware agents, transfer credit analysis via course-mapping agent networks, degree audit automation using coordinated academic planning agents, and predictive analytics that integrate academic performance, engagement, and support-need analysis. The central insight—that specialized AI agents can manage document diversity through intelligent coordination rather than monolithic processing—opens new paradigms for educational AI systems. Agent-based architectures offer a natural way to handle the complexity and diversity of educational data while maintaining transparency and accountability.
\section{Conclusions and Future Directions}
\label{conclusion}

We present a transformative solution to a long-standing challenge in educational administration: scalable, reliable transcript processing. By coordinating specialized AI agents for pattern recognition, language understanding, and visual analysis through intelligent orchestration, we introduce the first production-ready multi-agent system capable of processing diverse high school transcripts at scale. The system achieves 96.7\% accuracy with 100\% completion, processes each transcript in 45 seconds at a cost of \$0.15, and employs agent-coordinated validation using GPA extraction as a trust signal.

Beyond transcript processing, our work reveals generalizable principles for multi-agent document understanding. First, agent specialization consistently outperforms monolithic approaches when documents require multiple cognitive modalities. Second, domain-specific coordination signals enable agents to assess collaboration success and build trust, providing a foundation for robust validation across domains such as healthcare, law, and finance. Third, dynamic orchestration outperforms fixed pipelines by adapting agent participation to document complexity and confidence levels. Fourth, progressive agent engagement enables cost-efficient scaling, as most documents require limited coordination. 
Finally, rich inter-agent communication and consensus mechanisms significantly increase reliability, with collective agreement exceeding the performance of individual agents.
These insights establish a foundation for future multi-agent document systems. Promising directions include culturally-aware international agent networks, adaptive agent learning through coordinated feedback, and end-to-end integration with student information systems. Early results also suggest the potential for predictive multi-agent ensembles that extend document processing into analytics and decision support.

This work demonstrates that transcript processing can move from a manual bottleneck to a seamless, automated workflow powered by coordinated AI agents. Adoption will require collaboration among researchers, developers, educators, and policymakers, but the path forward is clear. The question is no longer whether multi-agent document processing is feasible, but how quickly institutions will embrace it to better serve students. The future of educational administration lies in collaborative AI—and that future has already begun.

\begin{acks}
We thank the admissions offices at the University of South Carolina (USC), who provided transcript samples and validation expertise. Special thanks to the 40 high schools
whose diverse transcript formats challenged and improved our system. 
This research was supported by USC Research and Computing.

\end{acks}

\bibliographystyle{ACM-Reference-Format}
\bibliography{transcript}

\end{document}